\documentclass[preprint,authoryear,12pt]{elsarticle}

\usepackage{verbatim}

\usepackage{geometry}

\usepackage{color,soul}

\usepackage[hidelinks]{hyperref}
\hypersetup{
  colorlinks   = true, 
  urlcolor     = blue,
  linkcolor    = blue, 
  citecolor    = blue 
}

\usepackage{graphicx}
\graphicspath{{./images/}}
\DeclareGraphicsExtensions{.pdf,.png,.jpg}
\usepackage{placeins}
\usepackage{caption}
\usepackage{subcaption}

\usepackage{booktabs}
\usepackage[table,xcdraw]{xcolor}
\usepackage{graphicx}
\usepackage{adjustbox}
\usepackage{multirow}

\usepackage{amssymb}
\usepackage{gensymb}
\usepackage[mathscr]{euscript}
\usepackage{bm}
\usepackage{mathtools}
\usepackage{nccmath}
\usepackage{amsmath}
\usepackage{braket}
\usepackage{dutchcal}

\usepackage{algorithm}
\usepackage{algpseudocode}

\usepackage{lineno, hyperref}

\usepackage{lipsum}

\journal{ISPRS Journal of Photogrammetry and Remote Sensing}
\begin{document}

\begin{frontmatter}

    \title{StructuredMesh: 3D Structured Optimization of Façade Components on Photogrammetric Mesh Models using Binary Integer Programming}

    \author{Libin Wang}
    \author{Han Hu\corref{cor1}}
    \author{Qisen Shang}
    \author{Bo Xu}
    \author{Qing Zhu}
    \cortext[cor1]{Corresponding Author: han.hu@swjtu.edu.cn}

    \address{Faculty of Geosciences and Environmental Engineering, Southwest Jiaotong University, Chengdu, China}
    \begin{abstract}
The lack of façade structures in photogrammetric mesh models renders them inadequate for meeting the demands of intricate applications. Moreover, these mesh models exhibit irregular surfaces with considerable geometric noise and texture quality imperfections, making the restoration of structures challenging. To address these shortcomings, we present StructuredMesh, a novel approach for reconstructing façade structures conforming to the regularity of buildings within photogrammetric mesh models. Our method involves capturing multi-view color and depth images of the building model using a virtual camera and employing a deep learning object detection pipeline to semi-automatically extract the bounding boxes of façade components such as windows, doors, and balconies from the color image. We then utilize the depth image to remap these boxes into 3D space, generating an initial façade layout. Leveraging architectural knowledge, we apply binary integer programming (BIP) to optimize the 3D layout's structure, encompassing the positions, orientations, and sizes of all components. The refined layout subsequently informs façade modeling through instance replacement. We conducted experiments utilizing building mesh models from three distinct datasets, demonstrating the adaptability, robustness, and noise resistance of our proposed methodology. Furthermore, our 3D layout evaluation metrics reveal that the optimized layout enhances precision, recall, and F-score by 6.5\%, 4.5\%, and 5.5\%, respectively, in comparison to the initial layout.
    \end{abstract}
    \begin{keyword}
        Oblique Photogrammetry \sep 3D Building Model \sep Binary Integer Programming \sep 3D Layout Regularization
    \end{keyword}
\end{frontmatter}

%\linenumbers

\section{Introduction}
\label{s:intro}

The recent advancements in Multi-View Stereo (MVS) technologies, coupled with the development of the penta-view oblique camera, have established oblique photogrammetry as the primary method for large-scale urban 3D reconstruction \citep{nan2017polyfit, fritsch2013oblique}. This technique provides automated and efficient acquisition of highly-detailed, photo-realistic, and large-scale photogrammetric mesh models \citep{gerke2016orientation, toschi2017oblique}. However, the mesh models do not possess crucial structured information, such as roofs, façades, and windows. This lack of structure limits the ability to support complex geospatial applications, and impedes the effective implementation of urban planning and disaster prevention and mitigation strategies \citep{wong2021modelling, vargas-munoz2021openstreetmap, biljecki2015applications}. Therefore, there is an urgent need to obtain 3D urban models that are enriched with critical structured information.

\begin{figure}[H]
    \centering
    \includegraphics[width=1.0\linewidth]{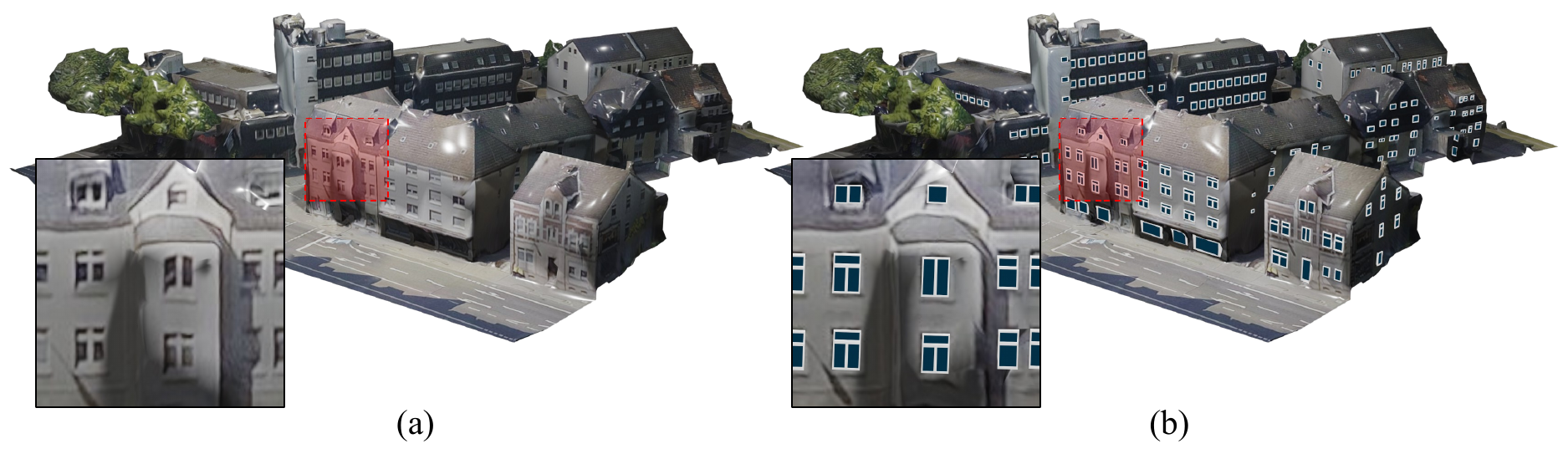}
    \caption{StructuredMesh. (a) The original photogrammetric mesh model lacks structured façade components, while (b) StructuredMesh achieves a comprehensive façade parsing of the entire building. This enables the reconstruction of specific and intricate façade components that adhere to the regularity of the building, such as equal relative elevation, equal size, and equal orientation.}
    \label{fig:structured_mesh}
\end{figure}

The CityGML standard \citep{groger2012citygml} defines three levels of detail (LOD) for the external structure of buildings, with LOD3 providing detailed geometric information about building façades. Research on automatic generation of LOD models has made significant progress, from traditional data-driven and model-driven approaches \citep{nan2017polyfit, verdie2015lod} to the latest end-to-end 3D reconstruction techniques based on deep learning \citep{yu2021automatic, li2022point2roof, chen2022reconstructing}. By combining these approaches with façade modeling techniques, highly detailed 3D models of buildings that are both regular and well-structured can be created \citep{verdie2015lod, kelly2017bigsur}. However, as the LOD model is a geometric abstraction of the real model, misalignment with the actual models is a challenge, necessitating substantial interaction for texture mapping and geometry reconstruction. This often requires sacrificing the advantages of authentic texture and automation via oblique photogrammetry. To address these challenges, we present StructuredMesh, which facilitates the direct reconstruction of façade structures incorporating the regularity of buildings within photogrammetric mesh models (Figure \ref{fig:structured_mesh}).

Nonetheless, implementing the StructuredMesh is not trivial, as two issues must be resolved.

\textit{1) Limitations of fragmented texture atlas and insufficient geometrical details}. The texture atlas of the photogrammetric mesh model exhibits discreteness and discontinuity \citep{zhu2021structureaware}, thereby presenting challenges for utilizing advanced deep learning models in object detection \citep{ren2015faster}. In addition, the insufficient level of geometric details on building façades poses a considerable challenge to accurately extracting façade components. This is attributed to the fundamental role of geometric characteristics, such as edges and corners, in the identification of object boundaries and the extraction of meaningful structured information.
\textit{2) Limitations due to noise and precision arise from the detected components}. As highlighted by \citet{zhu2021structureaware}, photogrammetric mesh models are prone to inherent texture quality issues, including blurring and distortion, which inevitably lead to a decline in the accuracy of identifying façade structures. Consequently, this loss of accuracy results in a deviation from the façade's regularity, ultimately impacting the aesthetic and geometric coherence of the 3D reconstruction.

In order to address the aforementioned issues, StructuredMesh employs a sophisticated binary integer programming (BIP) approach to tackle geometric misalignment in 3D space. Specifically, the study leverages multi-view rendering of building models captured by virtual cameras \citep{zhu2020leveraging}, in conjunction with state-of-the-art deep learning object detection techniques \citep{ren2015faster}, to semi-automatically extract bounding boxes of façade components, such as windows, doors, and balconies.
StructuredMesh then utilizes BIP to model the logical constraints, which include equal height, equal width, and geometrical alignment and applies global optimization to all the façades to ensure the geometric consistency of components in 3D space.
The detected components are then replaced by instances from the model library, which are scaled, translated and rotated with refined transformation through BIP. Finally, the instances can be stitched to the original photogrammetric mesh models to enhance the façade structure.

In summary, this article proposes two key innovations: 1) the detection of 3D layout of façade components on photogrammetric mesh models through the rendered color and depth image, and 2) a BIP-based method for optimizing the geometric regularity of the façade components. The paper is organized as follows: Section \ref{sec:related_work} provides an introduction to related works, while Section \ref{sec:approach} presents the methods in detail. Section \ref{sec:experiments} elaborates on the experimental data and results, and Section \ref{sec:conclusion} concludes the paper with our findings.

\section{Related works}
\label{sec:related_work}

We will provide a brief review by focusing on two aspects most closely related to our research, namely 1) image-based façade modeling and 2) binary integer programming for structure optimization.

\subsection{Image-based façade modeling}

The task of image-based façade modeling can be broadly classified into two categories: model-driven and data-driven approaches. Model-driven methods typically employ grammar-based approaches to search for combinations from a predefined set of grammars while fitting relevant parameters \citep{teboul2012parsing}. In principle, any combination of grammars and parameters can be utilized to decompose a given façade. However, optimizing these models using random sampling techniques such as Markov Random Fields or Markov Chain Monte Carlo can be a time-intensive and non-convergent process, particularly when the search space is vast \citep{koutsourakis2009single,kozinski2014image,kozinski2015mrf,tylecek2012stochastic}. Consequently, these methods often restrict the available grammar types and parameter values to those that conform to the associated rules, with manual derivation of grammars being necessary for decomposing complex façades. To overcome the limitations of model-driven approaches, some studies have investigated supervised learning methods by defining a limited number of common grammars and learning universal grammars from annotated façades and corresponding grammar parsing trees \citep{dehbi2017statistical,gadde2016learning}. However, the probability models generated from such methods are constrained by the training set and may be challenging to extend to general façades.

The data-driven approach to building façade segmentation is also exploited by a plethora of works. Unsupervised clustering methods do not require prior information. They typically divide building façade images into regular grids, assuming that the edges of objects such as windows and doors are parallel to the grid lines. The segmentation is based on low-level features of the image \citep{datta2008image,recky2010windows}, and clustering strategies are designed to determine specific regions formed by intersecting grid lines, such as using horizontal and vertical projections of gray-level histograms in the image, with the window edges determined by the grid lines corresponding to the peak values \citep{meixner2013interpretation}. The reliability of such methods is based on numerous conditions, such as the assumption that windows are arranged in a simple and regular pattern on the façade and that there is a significant difference in color between the windows and the walls, limiting the applicability. Machine learning methods train classification functions using prior data and then automatically segment unannotated data. Typical works include support vector machines, random forests, and decision trees, combined with low-level image features, their composite forms, or context information of pixels for façade segmentation \citep{frohlich2013semantic,gadde2018efficient,haugeard2009extraction}. However, such methods require elegantly-engineered image feature descriptors, making it difficult to ensure the reliability of the features constructed under different conditions, such as in scenes, lighting, and viewing angles.

With the improvement of hardware capabilities, target detection and semantic segmentation methods based on deep learning have achieved impressive progress in recent years. Deep convolutional neural networks \citep{krizhevsky2017imagenet} can directly learn high-level features from images. Such networks or their variants \citep{badrinarayanan2017segnet,he2017mask,ren2015faster} have gradually been applied to building façade segmentation and have achieved good results \citep{femiani2018facade,liu2020deepfacade,nishida2018procedural,wang2010approximate}. However, such methods usually focus more on network structures or training strategies, and how to use prior knowledge of façade regularities to guide network learning is still a challenging problem \citep{liu2020deepfacade,zhang2022rfcnet}.

While the model-driven approach can analyze a façade as a hierarchical structure, it is only feasible for examining façades with discernible rules due to its strict grammatical structure. For asymmetrical and misaligned façades, methods that account for weak architectural principles tend to be more adaptable \citep{mathias2016atlas}. Such methods lack clear grammatical definitions and usually represent windows and doors through their outer bounding boxes, which are then integrated into a façade layout. General constraints are applied to regularize the layout \citep{cohen2017symmetryaware,hensel2019facade,hu2020fast, jiang2016automatic, zhang2022rfcnet}. Nonetheless, the above method is only suitable for single façade images that have undergone ortho-rectification and cannot factor in the regularity of multiple façade layouts.

\subsection{Binary Integer Programming for Structure Optimization}

Binary integer programming is a mathematical optimization technique in which a linear objective function is minimized or maximized subject to a set of constraints. The decision variables are required to take on binary values of either 0 or 1. It is extensively employed in practical applications, including the reconstruction of LOD models \citep{li2016manhattan,monszpart2015rapter,fang2020connect,bouzas2020structure}, the final 3D model is composed of a combination of primitive elements. To accomplish this, primitive elements such as lines, planes, and blocks are extracted from point clouds or triangle meshes, and a binary variable is assigned to each primitive to indicate whether it is part of the final model. An energy function is then constructed to optimize the combination of primitive elements, which facilitates the reconstruction of the LOD model. Façade parsing \citep{jiang2015automatic,kozinski2015beyond} and other tasks also apply similar ideas. Building on the foundations of these excellent works, we utilized BIP to optimize the 3D layout of building façades. Moreover, our method goes beyond previous approaches by fully leveraging the powerful logical operation rules \citep{williams2009integer} inherent to BIP.

\section{Methods}
\label{sec:approach}

\subsection{Overview and problem setup}
\label{ssec:overview}

\subsubsection{Overview of the approach}

\begin{figure}[H]
    \centering
    \includegraphics[width=1.0\linewidth]{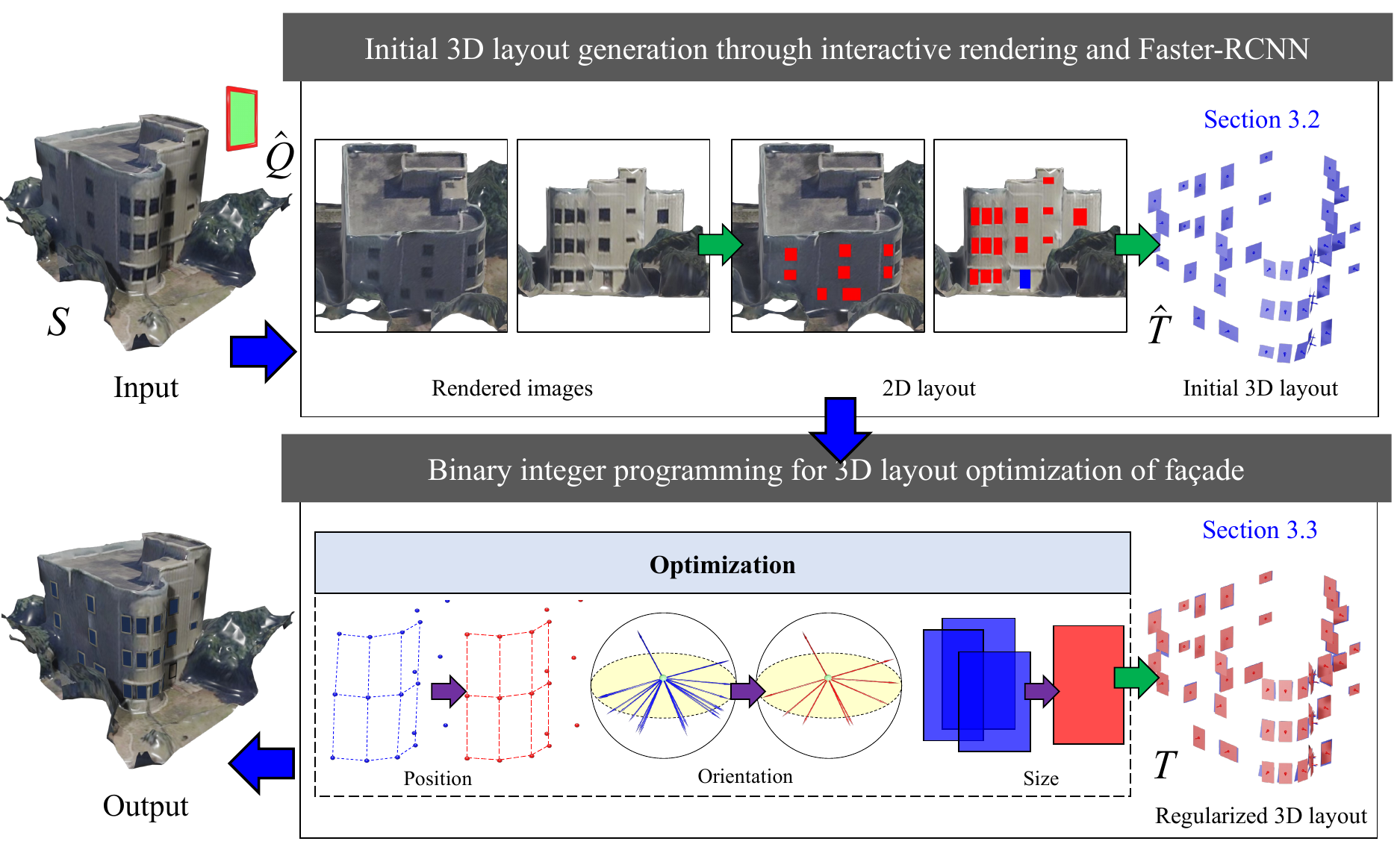}
    \caption{Overview of the workflow. The two parts are detailed in the following subsections.}
    \label{fig:overview}
\end{figure}

The overall pipeline, as depicted in Figure \ref{fig:overview}, consists of two distinct components. Firstly, we use a virtual camera to capture multi-view façade texture images in orthographic projection. We then employ the Faster-RCNN \citep{ren2015faster} to semi-automatically extract the bounding boxes of components and associate independent façade instances in the model library. Subsequently, we convert the bounding boxes to 3D space to obtain the initial 3D layout, after which we optimize the position, orientation, and size of each component in the layout, utilizing both architectural knowledge and the BIP algorithm. Finally, we stitch the façade components with the photogrammetric mesh models to obtain a structured mesh model.

\subsubsection{Problem setup}
\label{sssec:problem_setup}

\textbf{Mesh models $S$ and components $Q$}. We typically concentrate on a specific portion of the photogrammetric mesh model, which represents a building, denoted as $S$. Additionally, we use a collection of pre-built façade components with unit sizes, represented as $\hat{Q}$, as inputs. A transformation matrix is applied for the final components $Q$ as below. As a building comprises numerous components, we stitch them sequentially, generating intermediate results $S_i$. Both $S$ and $Q$ are considered as manifold meshes without topological errors, such as self-intersections or stray triangles.

\textbf{Transformation matrix of components $T$}. To determine the initial position of each component $Q$, we employ object detection techniques on the texture of $S$ (see Subsection \ref{ssec:init_layout}). We further optimize the placement of these components through BIP in our work (see Subsection \ref{ssec:bip_model}), obtaining the spatial transformation $T\in\mathbb{R}^{4\times4}$ for the components as $Q=T\hat{Q}$.

\textbf{Geometric parameters of components, $p$, $z$, $w$, $h$ and $n$}. It should be noted that the transformation matrix $T$ is not directly optimized in BIP but the decomposed geometric parameters are. The horizontal coordinates $p$, relative elevation $z$, width $w$, height $h$ and orientation $n$ are considered. The mapping between all the parameters and transformation mapping is invertable.

\subsection{Initial 3D layout generation through interactive rendering and Faster-RCNN}
\label{ssec:init_layout}

\subsubsection{Generation of color and depth maps through interactive rendering.}

We present a method for interactive selection of the initial components from the subset building mesh, wherein orthogonally rendered images are captured to detect façade primitives. As the building model lacks structured information and may contain noise, obtaining an optimal rendering plane parallel to the façade can be arduous. Therefore, we do not impose strict constraints on the pose of the scene camera. The graphics rendering pipeline efficiently projects a photogrammetric mesh model onto the screen space, displaying its texture information. To achieve this transformation, the pipeline requires the projection matrix and view matrix. We utilize the OpenSceneGraph's ortho routine \citep{osg} to construct these two matrices. The parameter values of the matrices are automatically determined by the minimum bounding box of the building model and the virtual camera's pose. The direct output of the rendering pipeline is a color image that accurately represents the texture of the building model. However, relying solely on the color image can only provide the 2D layout of the building. Therefore, we additionally create a frame buffer with the same size as the color image to store the depth information of the scene from the virtual camera's viewpoint, which is the depth image, as previously reported \citep{zhu2020leveraging}. It is worth mentioning that through the depth image, every point in the color image can be accurately mapped to the 3D model. With these tools at our disposal, we can generate the 3D layout of the building model from the 2D layout, as thoroughly explained in Subsection \ref{ssec:faster_rcnn}.

\subsubsection{Detection of initial components using Faster-RCNN}
\label{ssec:faster_rcnn}

\begin{figure}[H]
    \centering
    \includegraphics[width=\linewidth]{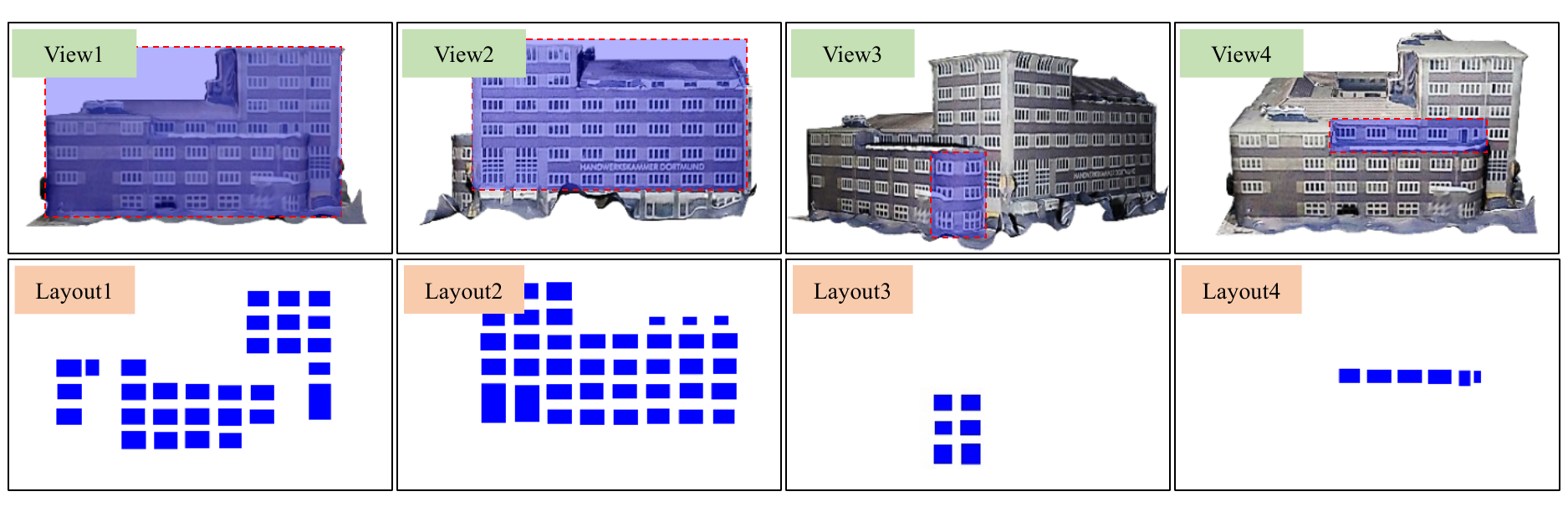}
    \caption{Instances of the interactive creation of rendered images and the detection of façade components.}
    \label{fig:2d_layout}
\end{figure}

\textbf{Detection of façade components}. We utilize Faster R-CNN \citep{ren2015faster} for the automatic extraction of objects in the rendered images, as detailed in our earlier work \citep{hu2020fast}. The training dataset consists of 606 building façade images from the commonly used open-source façade dataset \citep{tylevcek2013spatial}, with 12 artificially labeled objects, where our focus is limited to windows, doors, and balconies. Furthermore, we obtain 600 façade images from experimental data, manually labeling them for training purposes. Despite strict orthorectification, the reliability of object extraction results may be affected by distortion, blurring, occlusion, and other defects in textures. 
Additionally, the orthorectification of non-planar façade images remains unresolved. Hence, we only select the region of interest in the rendered image under each viewing angle and use it directly as the pipeline input. Our method achieves the accuracy, precision, recall, and F-score of 84.8\%, 77.2\%, 65.9\%, and 0.711, respectively, indicating acceptable results. By using manual sketching as an aid, the extraction of façade components can be efficiently accomplished. Figure \ref{fig:2d_layout} depicts some examples of the final 2D façade layout under various views.

\begin{figure}[H]
    \centering
    \includegraphics[width=1.0\linewidth]{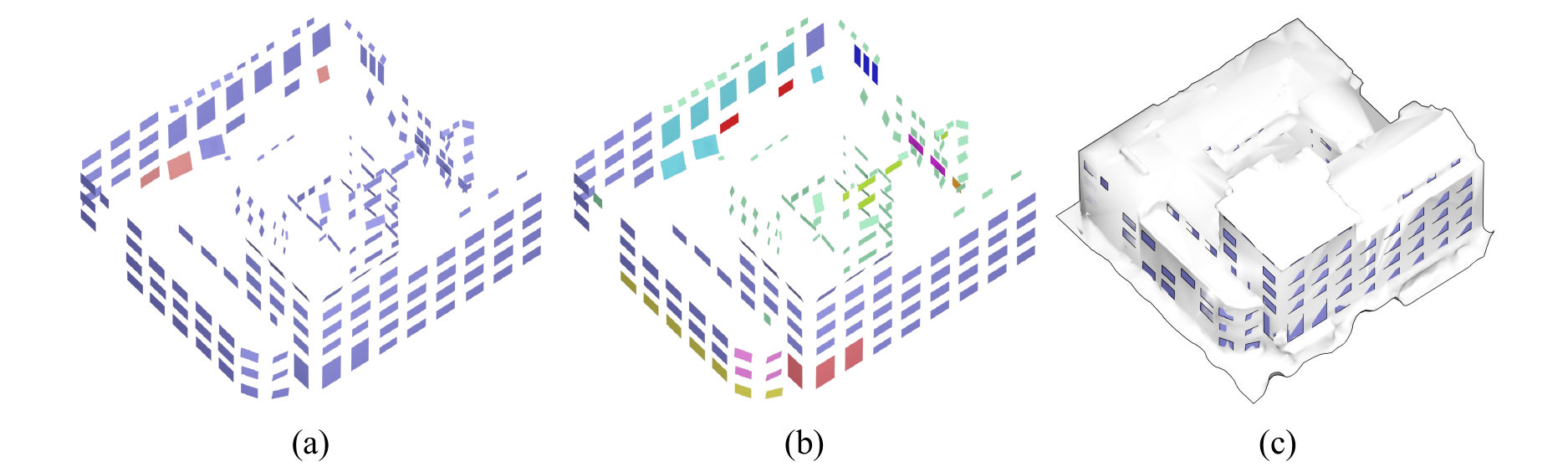}
    \caption{Generation of the initial 3D layout. (a) Different object categories; (b) Different types of selected models in the library; (c) Overlay of the mesh and initial layout.}
    \label{fig:init_3d_layout}
\end{figure}

\textbf{Generation of initial transformation matrix $\hat{T}$}. To generate the initial transformation $\hat{T}$ of each component $Q$, we leverage the depth maps produced from the rendering process discussed earlier. By taking into account the position and size of the bounding boxes on the 2D layout and the camera's pose, we derive the initial 3D layout through projection transformation. As illustrated in Figure \ref{fig:init_3d_layout}a, distinct colors indicate different object categories, including windows, doors, and balconies. We also establish an interactive association relationship between each bounding box and the corresponding 3D component in the model library, as highlighted by the different colors in Figure \ref{fig:init_3d_layout}b. Using orthogonal projection, we can deduce the widths and heights of the components in 3D Euclidean space from the projection's scale and the detected object bounding boxes' sizes. We then determine the position and orientation of the components by fitting the four corners back-projected to 3D space based on the depth map. The initial transformation matrix $\hat{T}$ is defined by the scale, position, and orientation \citep{pauly2008discovering}. Subsequently, Figure \ref{fig:init_3d_layout}c portrays the transformed objects $\hat{T}\hat{Q}$ overlaid on the building surface, which exhibit certain deviations in position, orientation, and size. Hence, optimization is essential to achieve better regularization.

\subsection{Binary integer programming for 3D layout optimization of façade}
\label{ssec:bip_model}

In our previous work \citep{hu2020fast}, we utilized BIP to optimize the 2D layout of a single façade image. This paper expands upon this methodology by extending the regularities to 3D space with multiple façades. To achieve regularized location, scale, and orientation for each component $Q$ encoded in the transformation matrix $T$, we present a logical optimization approach. The alignment constraints, such as aligned elevations for location, equal widths for sizes, and parallel orientation, are modeled using BIP, as described in Subsection \ref{ssec:obj_func} and Subsection \ref{ssec:bip_constraints}. The modeling details and optimization techniques are outlined in Subsection \ref{ssec:bip_parameter} and Subsection \ref{ssec:bip_optimization}.

\textbf{Basic logical operation}. The fundamental logical operations, including and ($\land$), or ($\vee$), not ($\neg$), xor ($\oplus$), etc., can be explicitly modeled using binary variables $x\in\{0,1\}$ with BIP, as shown in Table \ref{tab:bip}.

\begin{table}[htbp]
\centering
\caption{Expression for the basic logical operation using BIP.}
\label{tab:bip}
\begin{tabular}{lllll}
\hline
Expression & $z = x \land y$ & $z = x \vee y$ & $z = \neg x$ & $z = x \oplus y$ \\ \hline
BIP encoding & \begin{tabular}[c]{@{}l@{}}$z \geq x + y - 1$\\ $z \leq x$\\ $z \leq y$\end{tabular} & 
\begin{tabular}[c]{@{}l@{}}$z \leq x + y$\\ $z \geq x$\\ $z \geq y$\end{tabular} & $z + x = 1$ & 
\begin{tabular}[c]{@{}l@{}}$z \leq x + y$\\ $z \geq x - y$\\ $z \geq y - x$\\ $z \leq 2 - x - y$\end{tabular} \\ \hline
\end{tabular}
\end{table}

\textbf{Attributes space $\mathbb{A}$}. In contrast to previous work \citep{hensel2019facade}, our approach formulates the optimization problem of the component's geometrical transformation as a selection problem, rather than directly optimizing the values. To accomplish this, we define an attribute space $\mathbb{A}$ for each parameter as described in Subsection \ref{sssec:problem_setup}.
\begin{equation}
    \mathbb{A}=\{A_1, A_2,\cdots, A_N\}
    \label{eq:att_space}
\end{equation}

\textbf{Selection vector $\xi$}. To facilitate the selection problem, we introduce a selection vector for each parameter,
\begin{equation}
    \xi_i^a = \{x_{i,1}^a,x_{i,2}^a,\cdots,x_{i,j}^a,\cdots,x_{i,N}^a\}
    \label{eq:sel_vec}
\end{equation}
 where the superscript $a\in\{p,z,w,h,o\}$ indicates the different attributes. And $x_{i,j}\in\{0,1\}$ denotes whether the $j$-th parameter in the model space $\mathbb{A}$ is chosen for the $i$-th component $Q_i$.

\textbf{The \textit{same} operator}. To enforce the constraint that the parameter for different components $Q_i$ and $Q_j$ should be the same, we introduce the $same(\xi_i,\xi_j)$ operator, which is modeled as follows.
\begin{equation}
same(\xi_{i},\xi_{j})=\neg ( (x_{i,1}\oplus x_{j,1}) \vee (x_{i,2}\oplus x_{j,2}) \vee \cdots \vee (x_{i,N}\oplus x_{j,N}) ) 
\end{equation}
The operator $same(\xi_{i},\xi_{j})$ evaluates to true (1) if the selection vectors $\xi_{i}$ and $\xi_{j}$ choose the same parameter values for all $N$ parameters, and false (0) otherwise.

\textbf{The \textit{enumeration} operator}. To regularize the layout, it is preferred that the number of selected types in the model space is sparse. To formalize this, we introduce the enumeration operator. Given a set of $m$ vectors $\Omega = \{\xi_{1},\xi_{2},\cdots,\xi_{m} \}$, the enumeration operator models the number of different types as follows.
\begin{equation}
enum(\Omega) = \left\| {\xi_{1} \vee \xi_{2} \vee \cdots \vee \xi_{m}} \right\|_{1}
\label{eq:enum}
\end{equation}
The operator $enum(\Omega)$ evaluates to the number of unique selection vectors in $\Omega$, i.e., the number of distinct component types chosen from the model space. By minimizing $enum(\Omega)$, we encourage the selection of a smaller number of distinct component types in the layout, leading to a more regularized design.

\subsubsection{Parameter space}
\label{ssec:bip_parameter}

\begin{figure}[H]
    \centering
    \includegraphics[width=1.0\linewidth]{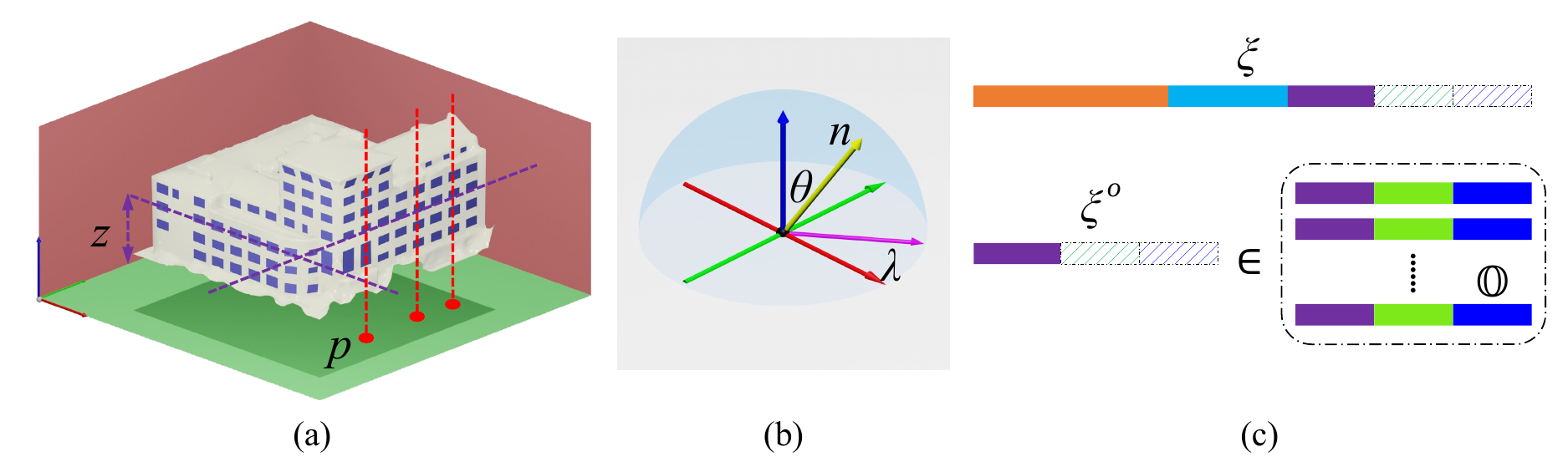}
    \caption{BIP Optimization Parameters for façade Regularization. The optimization parameters for the BIP in façade regularization are classified into two parts: (a) location, which includes the horizontal coordinates $p$ and relative elevation $z$; (b) orientation, which includes the normal vector $n$, horizontal angle $\lambda$, and elevation angle $\theta$. The final selection vector (c) $\xi$ for a single component is a concatenation of them. It's important to note that the set of three parameters in the orientation part are interdependent and therefore share the same values in the model space $\mathbb{O}$.}
    \label{fig:anotation}
\end{figure}

\textbf{Parameters $\xi^a$}. The optimization parameters for the BIP are depicted in Figure \ref{fig:anotation}. They comprise the horizontal position $p$, relative elevation $z$, sizes $w$ and $h$, normal vector of the plane $n$, horizontal angle $\lambda$, and elevation angle $\theta$ of the normal vector. The selection vector for a single component $Q$ is the direct concatenation of all the aforementioned parameters.
\begin{equation}
    \xi=(\xi^p,\xi^z, \xi^w, \xi^h, \xi^n,\xi^\lambda,\xi^\theta)
    \label{eq:parameter}
\end{equation}
The superscript denotes the different portions of the parameters.

\textbf{Model space clustering $\mathbb{P}$, $\mathbb{Z}$, $\mathbb{W}$, $\mathbb{H}$ and $\mathbb{O}$}. To construct the model space, we begin by computing the initial parameter value (Equation \ref{eq:parameter}) from the initial transformation $\hat{T}$. However, the presence of model noise can cause small numerical deviations, leading to a multitude of unnecessary solutions in the solution space, resulting in a considerable decline in efficiency. To mitigate this issue, we adopt an adaptive threshold to group the values obtained from the initial transformation (Subsection \ref{ssec:faster_rcnn}). Initially, we estimate the average absolute differences of the most adjacent components, designated as $\delta(\delta_p, \delta_z, \delta_n, \delta_w, \delta_h)$. We subsequently use a straightforward clustering algorithm \citep{ester1996density} with a threshold of $2\delta$ to create the model space for position $\mathbb{P}$, elevation $\mathbb{Z}$, orientation $\mathbb{O}$, width $\mathbb{W}$, and height $\mathbb{H}$.

\subsubsection{Objective functions}
\label{ssec:obj_func}

The objective function is the desire that should be minimized during the optimization of BIP. Specifically, two terms are considered, e.g., the data term $O_D$ and the regularization term $O_R$.  Therefore, the objective function for the BIP optimization consider both of them as below.
\begin{equation}
\underset{\xi}{\mathit{\min}}\quad{O_D(\xi) + O_R(\xi)}
\label{eq:o}
\end{equation}

\textbf{Data term}. 
The overall deviation between the optimized layout and the initial layout should not be too large. Therefore, we calculate $O_D$ as the difference between them \citep{hu2020fast},
\begin{equation}
    \epsilon^a_i=(\hat{a_i}-A_0,\hat{a_i}-A_1,\cdots,\hat{a_i}-A_{|\mathbb{A}|})^T
\end{equation}
where $a\in\{p,z,w,h,o\}$ and $i$ denote each attribute and component index, respectively. And $\hat{a_i}$ represents the value from the initial transformation $\hat{T_i}$ and $A_j$ is the value in the attribute space (Equation \ref{eq:att_space}).  In this way,  the total data term for all attributes and all elements can be briefly expressed as below.
\begin{equation}
    O_D=\sum_{a,i}{|\epsilon_i^a|\cdot\xi_i^a}
    \label{eq:odata}
\end{equation}
It should be noted that for each component $i$, only a single element in $\xi_i$ is selected as 1, e.g., only one element counts for each attribute $a$ and each component $i$.

\textbf{Regularization term}. Secondly, although the pose categories of façade components on the same building are quite limited, we still found that the Manhattan assumption \citep{coughlan2000manhattan} for real-world urban environment too restricted.  Therefore, StructuredMesh adopts the enumeration operator (Equation \ref{eq:enum}) to softly reduce the number of categories,
\begin{equation}
    O_R=\sum_{a}\omega^a\cdot enum(\Omega^a)
    \label{eq:oreg}
\end{equation}
where the weights $\omega$ is a weight to control the order of magnitude and $\Omega^a$ indicates all the vectors for a specific attribute (Equation \ref{eq:enum}).

\subsubsection{Modeling constraints}
\label{ssec:bip_constraints}

BIP can involve some linear equality constraints during the optimization. StructuredMesh exploits such constraints in two threads, i.e., feasibility and regularity. The former constraints the optimization to produce results aligned with certain priori knowledge and the latter enforces the tidiness of the structure.

\textbf{Regularity constraints}. In urban environment, buildings with different offsets for varying floor are common. And the noises of the photogrammetric meshes often impede the successful fusion if the orientation of the components $Q$ deviates too much from the mesh models. Fortunately, we have identified two effective constraints that can greatly enhance the regularity of the layouts. (1) For the same floor (the same $z$), the elevation angle $\theta$ should be the same; and (2) for the same column (the same $p$), the horizontal angle $\lambda$ should be the same. Thus, we introduce the following two regularity constraints.
\begin{equation}
    \begin{aligned}
        R1: & same(\xi_i^z,\xi_j^z)=same(\xi_i^{z \cup \theta},\xi_j^{z \cup \theta}) \quad \forall i,j \\
        R2: & same(\xi_i^p,\xi_j^p)=same(\xi_i^{p \cup \lambda},\xi_j^{p \cup \lambda}) \quad \forall i,j
    \end{aligned}
\end{equation}

\textbf{Feasibility constraints}. First of all,  a single value $x_{i,j}^a$ in the selection vector $\xi_i^a$ (Equation \ref{eq:sel_vec}) should be and must be selected.
\begin{equation}
    C_1: |\xi_i^a|=1\quad \forall i,a
\end{equation}
Secondly, different components $Q_i$  and $Q_j$ can not coincide with each other. Namely, they can not chose the same category for locations, i.e., horizontal location $p$  and relative elevation $z$. This constraint is enforced with the \textit{same} operator as below.
\begin{equation}
    C2: same(\xi_i^{p\cup z},\xi_j^{p \cup z})=0\quad \forall i\neq j
\end{equation}
Thirdly, as noticed in Figure \ref{fig:anotation}, the normal vector $n$, horizontal angle $\lambda$ and elevation angle $\theta$ should chosen the same set in the attribute space $\mathbb{O}$, i.e., attribute $o=n \cup \lambda \cup \theta$ . This is effectively modelled the constraints below.
\begin{equation}
    C3: \sum_{\xi_{j}^{o} \in \mathbb{O}}{same\left( {\xi_{i}^{o},\xi_{j}^{o}} \right) = 1} \quad \forall i
\end{equation}

\subsubsection{Optimization}
\label{ssec:bip_optimization}

The integer programming is notoriously difficult for optimization. We use Gruobi \citep{gurobi}, which is widely regarded as one of the most powerful and efficient optimization solvers, to solve our BIP model. Besides, a simple trick has been found to emperically accelerate the optmization speed. Because each independent variable of component $i$ has a certain range, that is, when the $k$ component $\varepsilon_{i,k}^{\gamma}$ of the residual vector $\varepsilon_{i}^{\gamma}$ exceeds the threshold $\delta^{\gamma}$, the $k$ component of the corresponding selection vector $
\xi_{i,k}^{\gamma} = 0$, where $\delta^{\gamma} = 5\Delta$. Because of the above strategy, unlike \citep{kelly2017bigsur}, the $same(\cdot)$ and $enum(\cdot)$ operators in this paper are not actually executed on the two full-length selection vectors $\xi_{i}$ and $\xi_{j}$, instead, calculations are performed only on dimensions where there is an unknown at the corresponding position in $\xi_{i}$  and $\xi_{j}$.

\section{Experimental evaluations and analyses}
\label{sec:experiments}

\subsection{Dataset description}
\label{ssec:dataset_description}

\begin{figure}[htbp]
    \centering
    \includegraphics[width=1.0\linewidth]{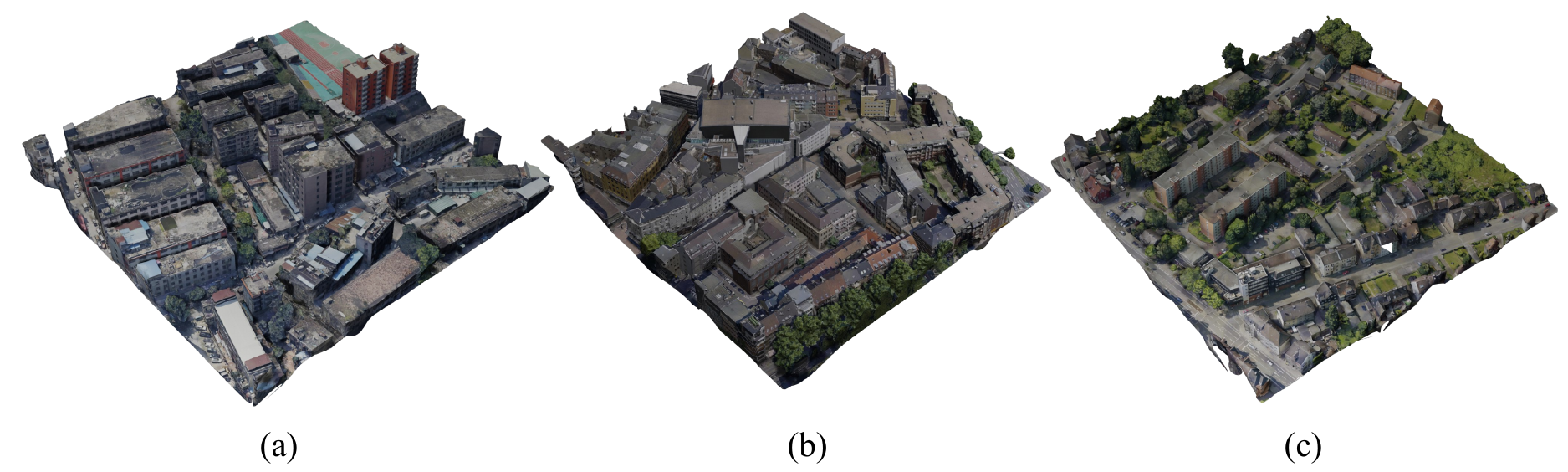}
    \caption{Datasets. (a) Shenzhen; (b) Dortmund; (c) Zeche Zollern.}
    \label{fig:datasets}
\end{figure}

\begin{figure}[htbp]
    \centering
    \includegraphics[width=1.0\linewidth]{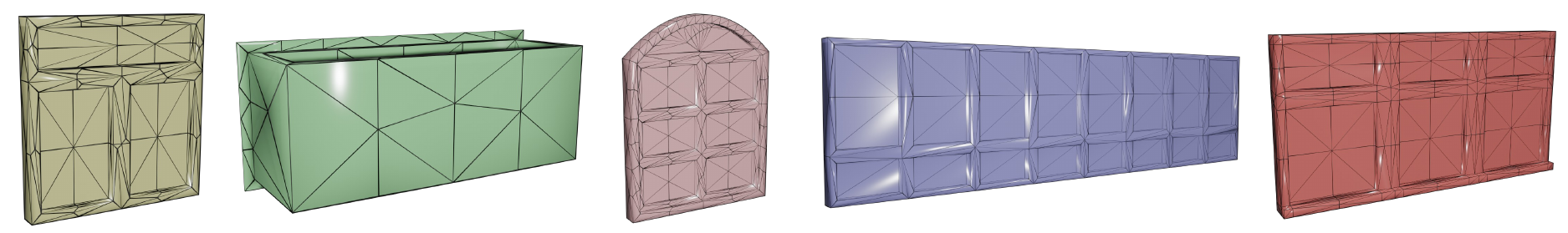}
    \caption{Façade components}
    \label{fig:components}
\end{figure}

In order to verify the effectiveness of our method, we use oblique images of three regions to experiment, which are the Shenzhen dataset collected independently, the Dortmund dataset and the Zeche Zollern dataset published by \citep{nex2015isprs}. The photogrammetric meshes obtained are shown in Figure \ref{fig:datasets}. The actual data processed in this paper is a single building or a piece of building clipped from them. Table \ref{tab:reg} gives the specific data source of the experimental results with the figure indexes. In addition, we pre-design several common components and store them in the model library, as show in Figure \ref{fig:components}.

\subsection{Qualitative experiment}
\label{ssec:qualitative_experiment}

\begin{figure}[H]
    \centering
    \includegraphics[width=1.0\linewidth]{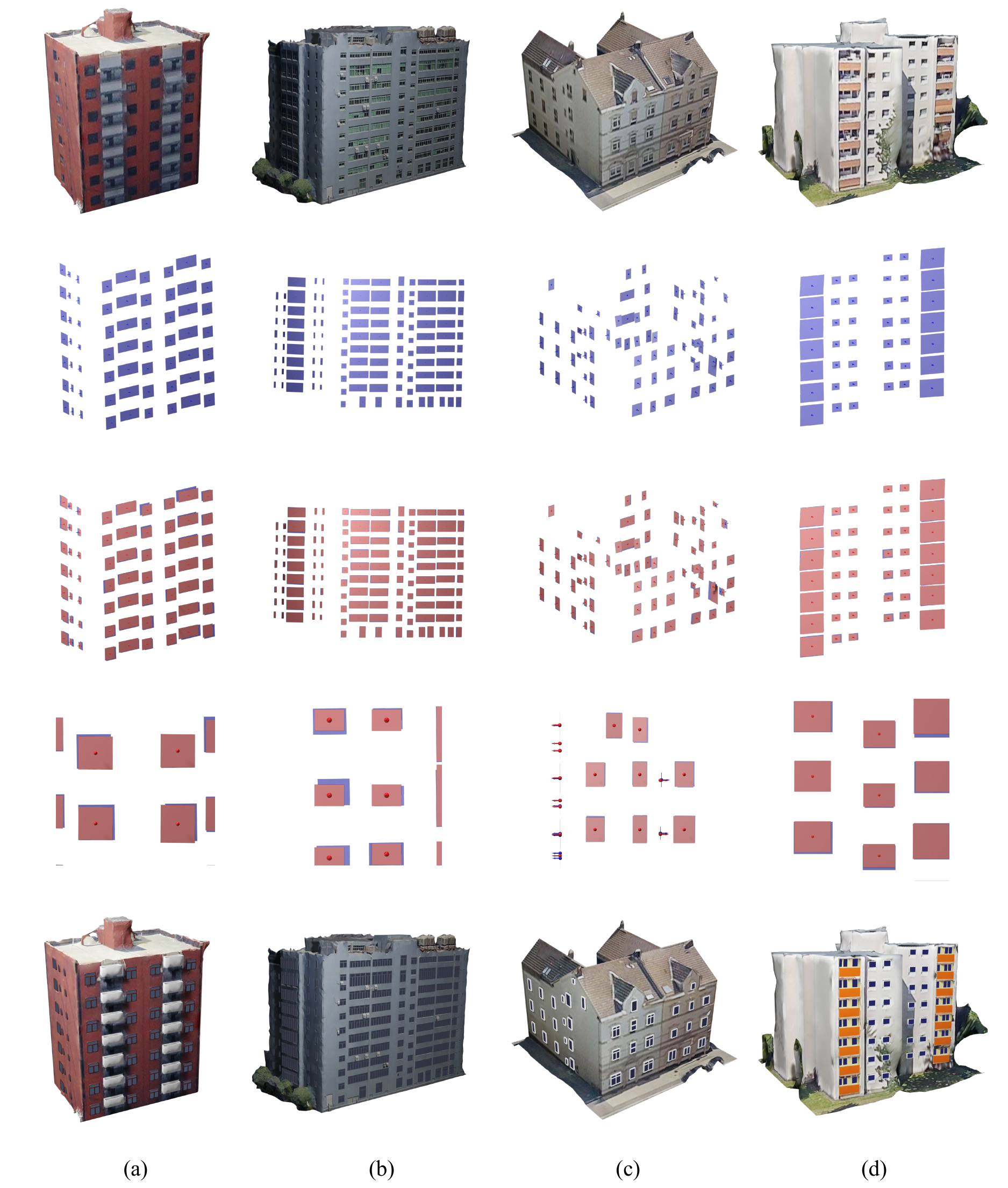}
    \caption{The results of our method for four simple building models are displayed in a top-to-bottom sequence, presenting the original models, followed by the initial 3D layouts, regularized 3D layouts overlaid on the initial layouts, the details, and the final models.}
    \label{fig:simple_models}
\end{figure}

The experimental results of four simple building models are depicted in Figure \ref{fig:simple_models}. Each row, from top to bottom, represents the original models, the initial 3D layouts, the regularized 3D layouts (overlaid on the initial 3D layouts), the details, and the final models. Figure \ref{fig:simple_models}a and  Figure \ref{fig:simple_models}b are derived from the Shenzhen dataset, which boasts a higher image resolution and a more pronounced contrast between façade objects and walls. However, texture distortion remains an issue, as a result, even components with the same geometric structure have different sizes in texture. While a simple method, such as utilizing the average size directly based on the prior knowledge from Figure \ref{fig:init_3d_layout}a and Figure \ref{fig:init_3d_layout}b, may produce more pleasing visual results, it could result in the overlapping of components that do not match the original model. Therefore, we do not strongly restrict the sizes of components. Figure \ref{fig:simple_models}c and  Figure \ref{fig:simple_models}d, taken from the Zeche Zollern dataset, exhibit lower image resolution and reconstructed models with blurry, occluded, and missing textures. Such situations are widespread in practical data. In our experiments, relying solely on the deep learning object detection pipeline failed to produce dependable initial layouts. As a result, we permit both interactively drawn bounding boxes and automatically obtained bounding boxes to be utilized as optimization targets, enhancing the method's practicality.

\begin{figure}[H]
    \centering
    \includegraphics[width=1.0\linewidth]{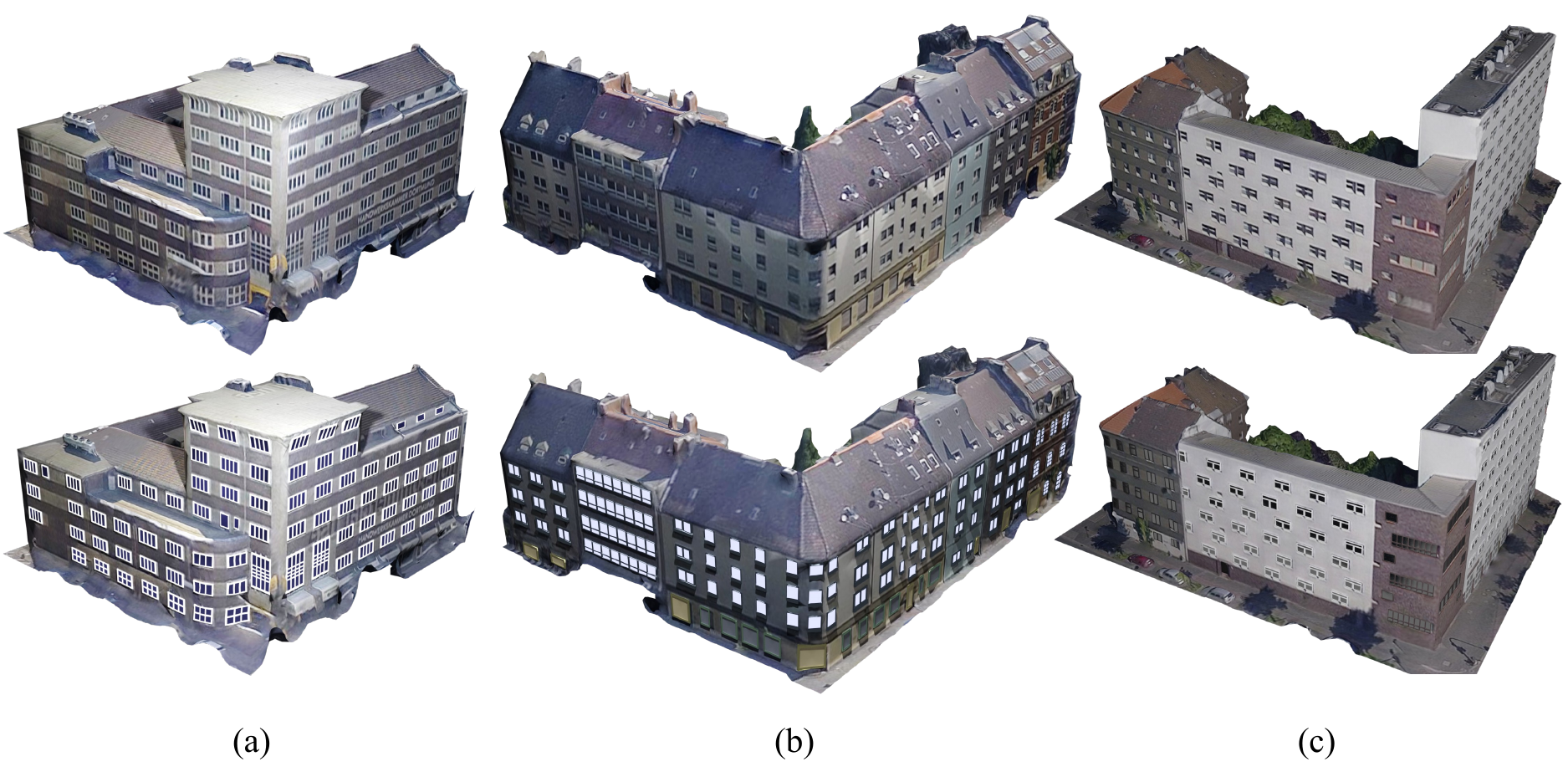}
    \caption{The experimental results for three complex building models are presented, with the original models displayed in the top row and the corresponding results obtained by our proposed method shown in the bottom row. These buildings are characterized by a high degree of complexity, featuring a large number of components and significant noise.}
    \label{fig:complex_models}
\end{figure}

\begin{figure}[H]
    \centering
    \includegraphics[width=1.0\linewidth]{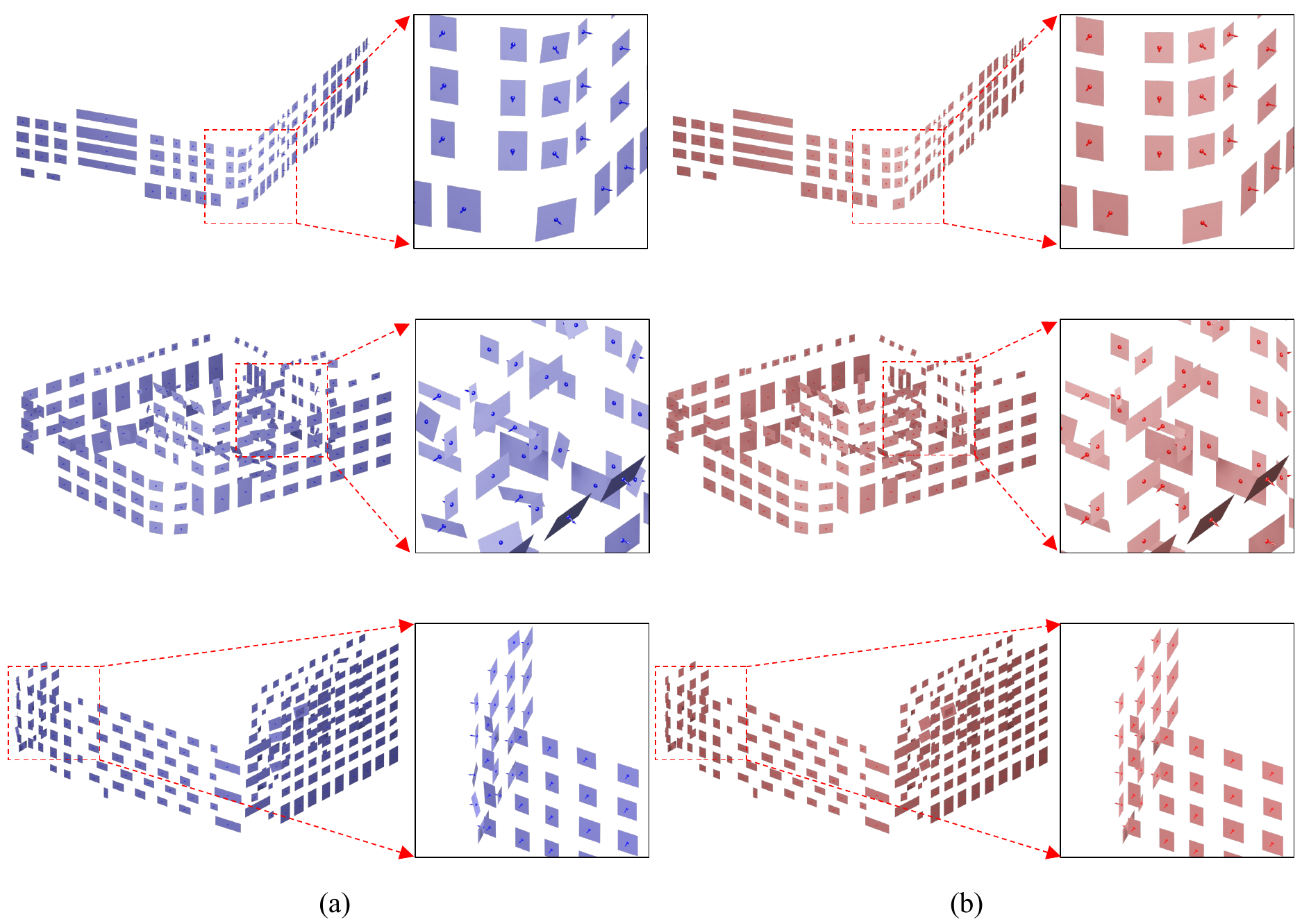}
    \caption{3D layouts of complex building models with local details. The comparison between (a) initial 3D layouts and (b) regularized 3D layouts demonstrates the effectiveness of our proposed method in eliminating minor differences and reducing misalignment of component poses caused by noise.}
    \label{fig:3d_reg}
\end{figure}

The experimental results of our proposed method applied to more complex building models are presented in Figure \ref{fig:complex_models}. The top row displays the original models, while the bottom row shows the corresponding experimental results. All three buildings are taken from the Dortmund dataset and are characterized by a large number of components and significant noise in the model. This noise is particularly evident on the unshown side of the models, as depicted in Figure \ref{fig:init_3d_layout}. Furthermore, the presence of non-planar surfaces, such as corners and eaves, where façade components are distributed, adds complexity to the optimization task. Figure \ref{fig:3d_reg} illustrates the initial (Figure \ref{fig:3d_reg}a) and optimized (Figure \ref{fig:3d_reg}b) layouts of these complex buildings, with local details presented on the right of each layout. Our method effectively eliminates small differences in the initial layout and reduces the misalignment of component poses caused by noise, demonstrating its robustness and effectiveness in handling complex building models.

\subsection{Quantitative experiment}
\label{ssec:quantitative_experiment}

It can be challenging to provide quantitative indicators for describing the regularity of building layouts. Generally, a layout appears more regular when it has fewer categories. To demonstrate this, we have counted the number of categories for the relevant building parameters before and after optimization and presented our findings in Table \ref{tab:reg}. The symbols used have been explained in Section \ref{ssec:bip_model}. It should be noted that our method  ensures the regularity of building layouts while reducing categories.

\begin{table}[]
\centering
\caption{The relevant parameters of the building models and their layouts.}
\label{tab:reg}
\resizebox{\textwidth}{!}{
\begin{tabular}{cccccccccccccc}
\hline
\multirow{2}{*}{\begin{tabular}[c]{@{}c@{}}Biulding\\ Indexes\end{tabular}} &
  \multirow{2}{*}{$N$} &
  \multicolumn{5}{c}{Before} &
  \multicolumn{5}{c}{After} &
  \multirow{2}{*}{\begin{tabular}[c]{@{}c@{}}Figure\\ Indexes\end{tabular}} &
  \multirow{2}{*}{\begin{tabular}[c]{@{}c@{}}Data\\ Source\end{tabular}} \\
 &
   &
  $|\mathbb{P}|$ &
  $|\mathbb{Z}|$ &
  $|\mathbb{N}|$ &
  $|\mathbb{W}|$ &
  $|\mathbb{H}|$ &
  $|\mathbb{P}|$ &
  $|\mathbb{Z}|$ &
  $|\mathbb{N}|$ &
  $|\mathbb{W}|$ &
  $|\mathbb{H}|$ &
   &
   \\ \hline
1 & 42  & 42  & 42  & 42  & 23  & 29  & 6   & 30 & 3  & 2  & 2  & \ref{fig:simple_models}d  & Zeche Zollern \\
2 & 63  & 63  & 62  & 56  & 39  & 45  & 10  & 21 & 2  & 3  & 3  & \ref{fig:simple_models}a  & Shenzhen      \\
3 & 71  & 71  & 71  & 71  & 38  & 46  & 33  & 21 & 6  & 8  & 8  & \ref{fig:simple_models}c  & Zeche Zollern \\
4 & 101 & 101 & 95  & 101 & 77  & 69  & 48  & 20 & 6  & 10 & 9  & \ref{fig:complex_models}b & Dortmund \\
5 & 128 & 128 & 80  & 128 & 82  & 43  & 24  & 41 & 4  & 8  & 6  & \ref{fig:simple_models}b  & Shenzhen \\
6 & 207 & 206 & 202 & 206 & 145 & 145 & 111 & 21 & 16 & 12 & 10 & \ref{fig:complex_models}a & Dortmund \\
7 & 221 & 221 & 219 & 210 & 122 & 92  & 68  & 34 & 12 & 18 & 22 & \ref{fig:complex_models}c & Dortmund \\ \hline
\end{tabular}}
\end{table}

\begin{figure}[H]
    \centering
    \includegraphics[width=1.0\linewidth]{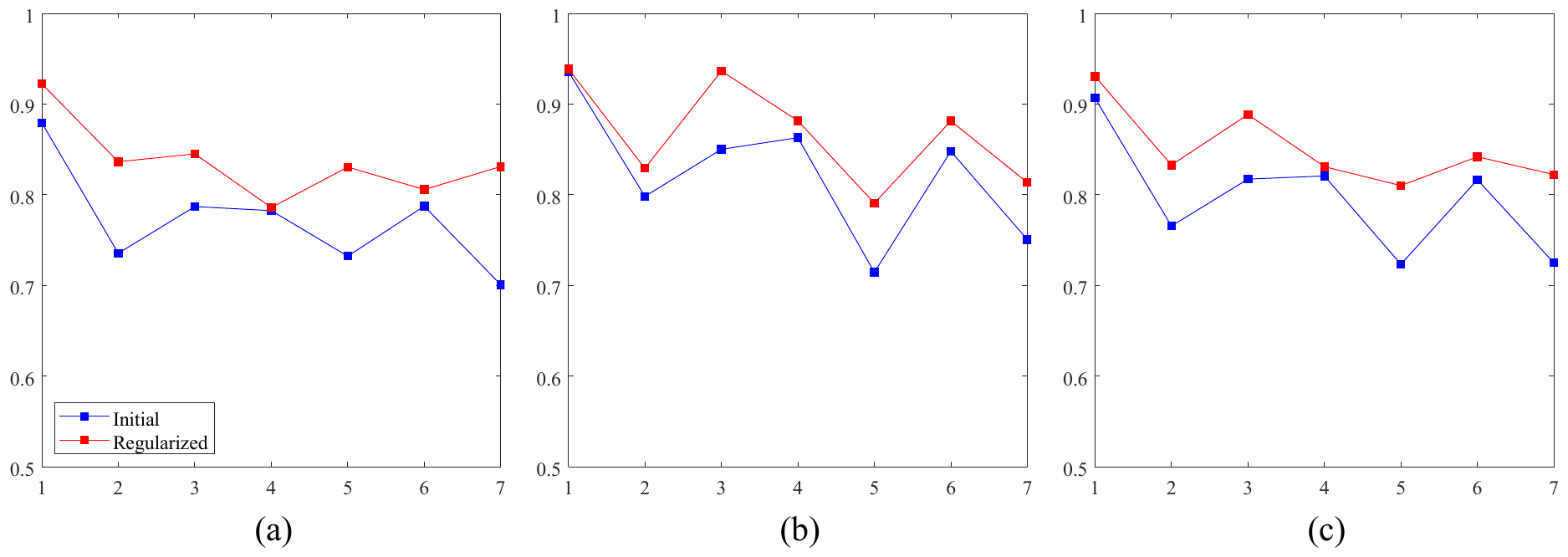}
    \caption{Evaluation metrics (a) precision, (b) recall, and (c) F-score of our method, with the true layout constructed interactively using 3D modeling software. The regularized layout demonstrated higher precision, recall, and F-score compared to the initial layout.}
    \label{fig:prf}
\end{figure}

We used precision ($P$), recall ($R$), and F-score ($F$) to evaluate our method, similar to other 2D layout evaluation metrics \citep{2016Automatic}. The calculation is shown in Equation \ref{eq:prf}, where we constructed the true layout $\overset{\sim}{L}$ interactively using 3D modeling software \citep{Blender} and $\Lambda_{L}$ is the sum of the areas of each polygon in layout $L$. We projected each polygon in $L$ onto the corresponding polygon in $\overset{\sim}{L}$ and used boolean intersection \citep{2009CGAL} to obtain a set containing multiple 2D polygons denoted as $L \cap \overset{\sim}{L}$. The relevant quantities were calculated for the initial and regularized layouts of the seven buildings mentioned above, and the results are illustrated in Figure \ref{fig:prf}. Compared to the initial layout, the regularized layout showed 6.5\%, 4.5\%, and 5.5\% higher average precision, recall, and F-score, respectively. Furthermore, as shown in Figure \ref{fig:prf}c, the overall evaluation metric F-score did not exhibit a significant pattern in the optimization effect for different building layouts. This is because the position, size, and orientation of each component all affect the value of this metric, and even if the orientation of certain polygons is corrected, it may not result in a significant improvement in the value of the overall layout (e.g., for building 4 and 6). Nevertheless, in general, the higher the F-score, the closer the layout is to the ground truth.
\begin{equation}
    \begin{aligned}
       & P = \Lambda_{L \cap \overset{\sim}{L}}/\Lambda_{L} \\
       & R = \Lambda_{L \cap \overset{\sim}{L}}/\Lambda_{\overset{\sim}{L}} \\
       & F = 2 \cdot P \cdot R/\left( {P + R} \right)
    \end{aligned}
\label{eq:prf}
\end{equation}

\subsection{Comparative experiment}
\label{ssec:comparative_experiment}
\begin{figure}[H]
    \centering
    \includegraphics[width=1.0\linewidth]{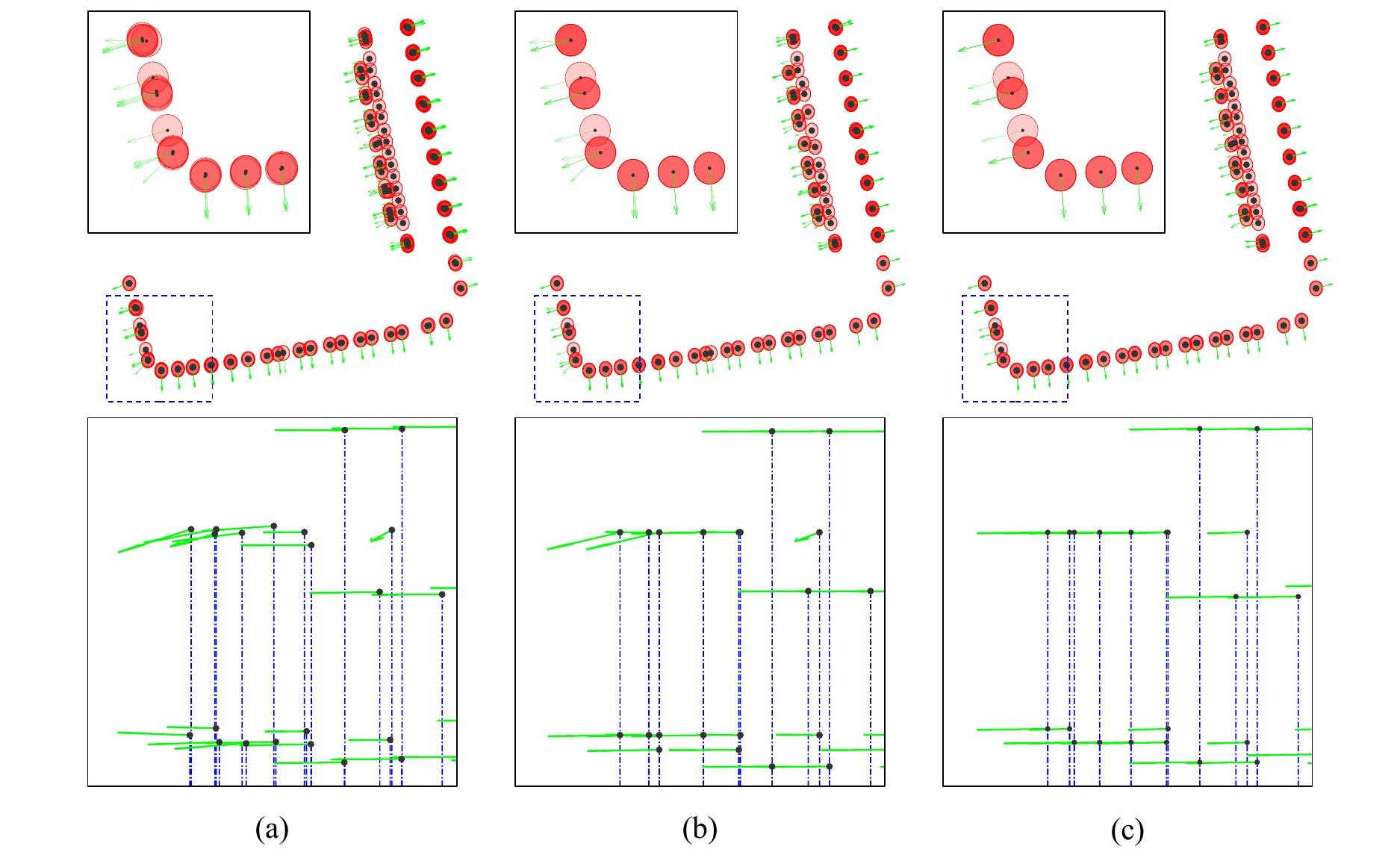}
    \caption{Comparison between the traditional clustering method and our approach. We utilized the adaptive clustering algorithm mean-shift to cluster the (a) raw data, resulting in (b), and compared it with the results of our approach in (c). The red circles represent the position in the top view, the green arrows represent the normal vectors, and the blue dotted lines depict the auxiliary lines.}
    \label{fig:constraint_comparison}
\end{figure}

There is currently a lack of research on optimizing 3D building façade layouts, but reducing the number of categories often leads to more regular designs, as noted by \citet{hensel2019facade}. In order to compare traditional clustering methods with our own, we employed mean-shift \citep{2002Mean,carreira2015review}, a method that can perform adaptive clustering without the need for pre-defined categories and has been widely used for optimizing geometric structures and façade layouts \citep{pauly2008discovering, kelly2017bigsur}. Figure \ref{fig:constraint_comparison} illustrates the results (position and orientation) of applying mean-shift to the building layout shown in Figure \ref{fig:complex_models}c, along with our own approach. The top row presents an overhead view of the entire layout and local details, while the bottom row shows the side view of local details. Figure \ref{fig:constraint_comparison}a depicts the raw data, Figure \ref{fig:constraint_comparison}b displays the outcomes of adaptive clustering using mean-shift, and Figure \ref{fig:constraint_comparison}c showcases the results of our approach. The findings indicate that mean-shift enhances regularity, although it is still affected by noise. Additionally, mean-shift was unable to prevent overlapping positions in 3D space in our evaluations.

\begin{figure}[H]
    \centering
    \includegraphics[width=1.0\linewidth]{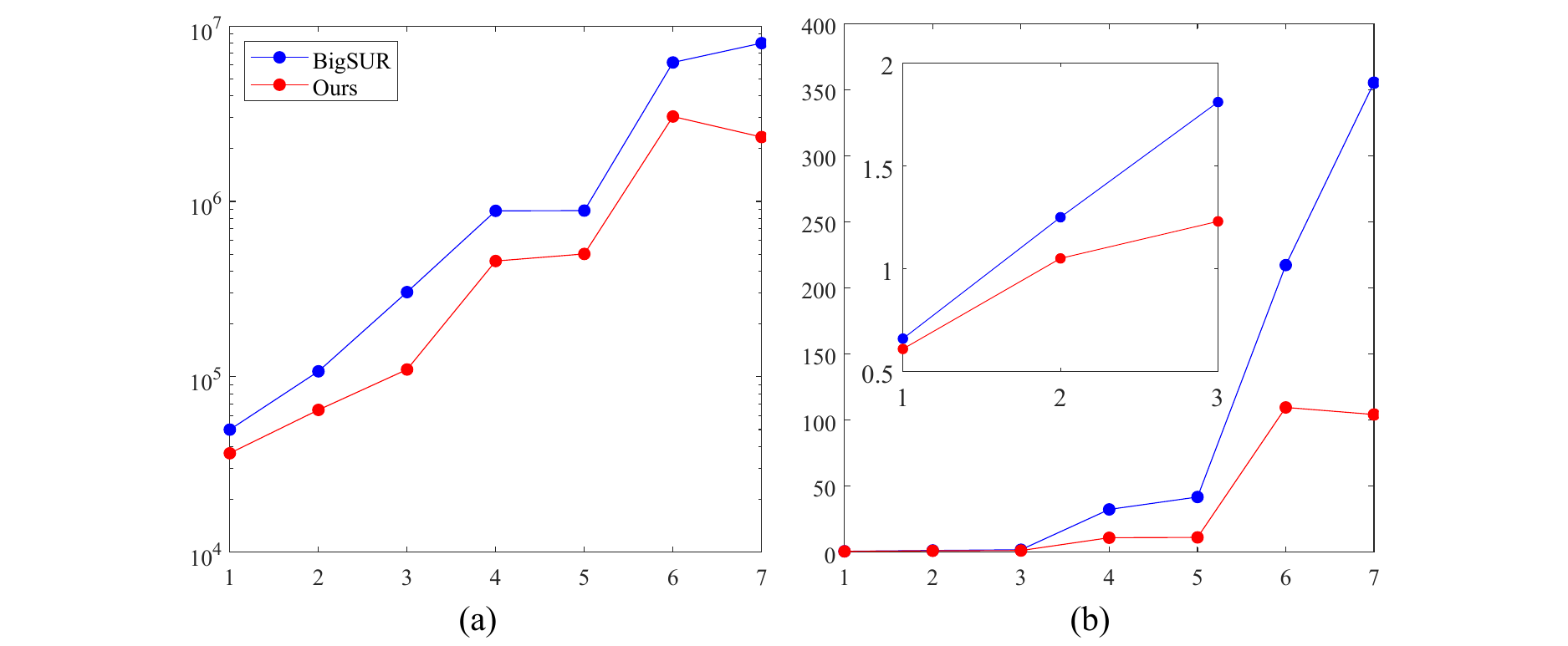}
    \caption{Efficiency comparison of our proposed optimization strategy and Bigsur's when optimizing the layouts of seven buildings. The comparison is based on the following relevant parameters: (a) Number of unknowns; and (b) Time consumption.} 
    \label{fig:reg_efficiency}
\end{figure}

BIP enables the creation of special algebraic operations using logical operations. However, it suffers from computational inefficiency due to its NP-hard nature, commonly referred to as combinatorial explosion. This issue becomes increasingly problematic as the number and types of building components grow. To address this challenge, we adopted the $same(\cdot)$ operation from Bigsur \citep{kelly2017bigsur} to construct our BIP model and developed related optimization strategies detailed in Section \ref{ssec:bip_optimization}. Figure \ref{fig:reg_efficiency} compares the key parameters of our strategy and Bigsur's in optimizing the layouts of the seven buildings mentioned above. Specifically, Figure \ref{fig:reg_efficiency}a illustrates the number of unknowns, and Figure \ref{fig:reg_efficiency}b shows the time consumption. Our proposed optimization strategy has considerably reduced the scale of the BIP model, with an average reduction of 50\%, resulting in a noteworthy decrease in memory usage. Furthermore, as the layout complexity increases, our approach leads to a significantly greater reduction in computation time, as illustrated in Figure \ref{fig:reg_efficiency}b.

\subsection{Robustness}
\label{ssec:robustness}

\begin{figure}[H]
    \centering
    \includegraphics[width=1.0\linewidth]{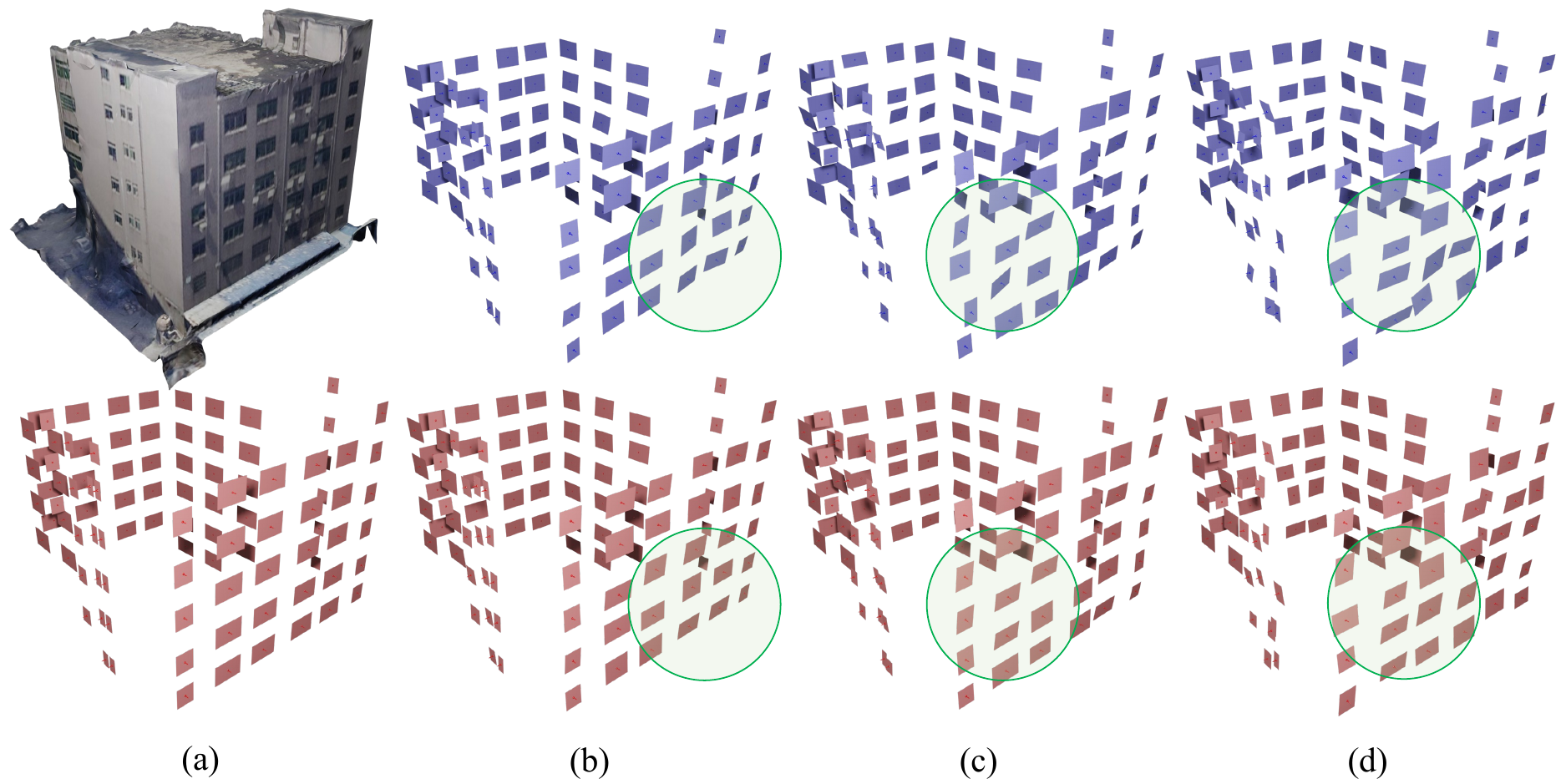}
    \caption{Robustness test. Incremental levels of Gaussian noise were introduced gradually to a true layout (a), and our method was employed to optimize them. The resulting layouts for variances of (b) $5\delta_{e}$, (c) $10\delta_{e}$, and (d) $15\delta_{e}$ are shown.}
    \label{fig:robustness_layout}
\end{figure}

\begin{figure}[H]
    \centering
    \includegraphics[width=0.5\linewidth]{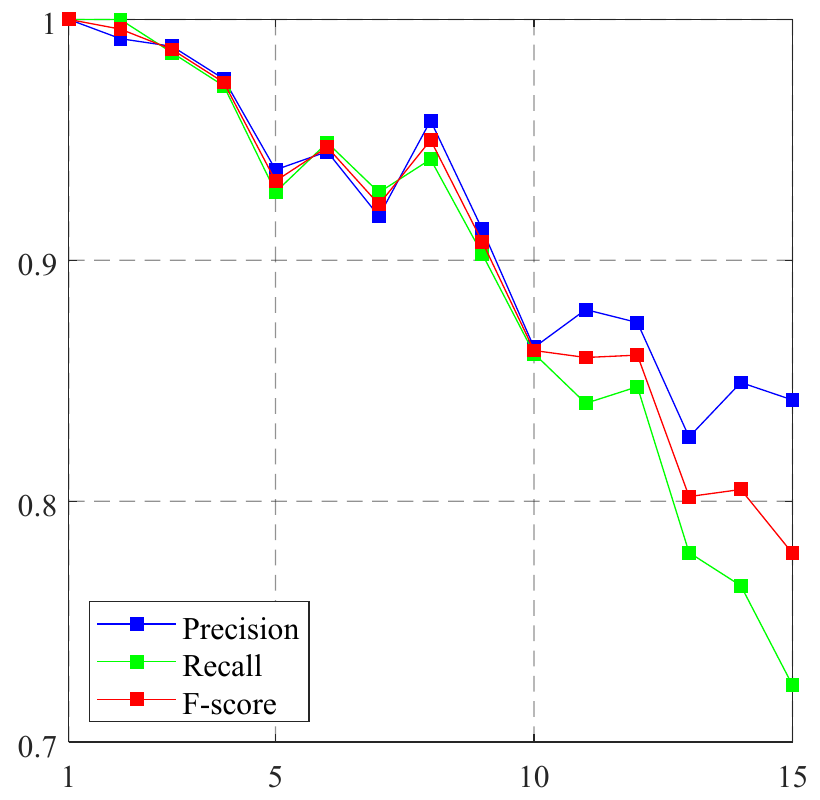}
    \caption{The quantitative metrics were evaluated with increasing levels of noise to assess the robustness of the proposed method.}
    \label{fig:robustness_prf}
\end{figure}

To evaluate the robustness of our approach, we gradually introduced Gaussian noise to a true layout, creating multiple initial layouts, and subsequently applied our method to optimize these layouts, followed by the application of Equation \ref{eq:prf} for quantitative assessment. The results are illustrated in Figure \ref{fig:robustness_layout}, where Figure \ref{fig:robustness_layout}a displays the original building and its true layout. Let $e$ be the length of the shortest edge of the polygons in the true layout, and let $\delta_{e}=0.005{e}$ be the base variance. Gaussian noise was gradually introduced to this layout with variances ranging from $\delta_{e}$ to $15\delta_{e}$ with a step size of $\delta_{e}$. The top row of Figure \ref{fig:robustness_layout}b, c and d presents the layouts subjected to Gaussian noise of $5\delta_{e}$, $10\delta_{e}$, and $15\delta_{e}$, respectively, while the bottom row displays the corresponding optimized layouts. Figure \ref{fig:robustness_layout} demonstrates that our method improves the position and orientation of the initial layout and exhibits a certain degree of denoising effect. The quantitative metrics with increasing noise are shown in Figure \ref{fig:robustness_prf}, and it can be observed that when the variance is less than $9\delta_{e}$, the F-score of our method remains above 0.9. As the noise level increases, the sizes of the polygons in the layout change significantly, leading to a noticeable fluctuation in the F-score.

\subsection{Discussion and limitations}
\label{sec:discussion}

To fully utilize the geometric and texture information of photogrammetric mesh models, we extracted and optimized the 3D layout of the façade components directly from the models. This approach improved the regularity and geometric details of the original façade, which has significant practical value.

However, there are still limitations to our method. For instance, we did not impose strict constraints on the sizes of the components due to the texture noise present in the photogrammetric mesh model. Even if a layout with relatively uniform sizes is obtained, some components may overlap partially. Presently, we have not addressed this issue, but we have ensured that the center points of the components do not overlap.

\section{Conclusion}
\label{sec:conclusion}

In this investigation, we unveil StructuredMesh, a sophisticated methodology that capitalizes on the logical operations of BIP to refine 3D geometric structures of façades. Distinctively, StructuredMesh eschews dependence on orthorectified façade images for modeling, exhibiting an exceptional capability for alleviating geometric noise. Furthermore, it embraces a comprehensive optimization strategy for multiple façades, ensuring seamless congruity among components within the 3D space. Intrinsically, our BIP model operates as a clustering technique, albeit with comparatively modest efficiency. Recently, the emergence of advanced deep learning methods for discerning geometric parameters and relationships through differentiable rendering \citep{Li:2020:DVG} and differentiable optimal transport \citep{cuturi2019differentiable} has been witnessed. These pioneering approaches manifest extraordinary potential to bridge the gaps between images and geometric components, heralding a new opportunity for the conjoint detection and optimization of façade geometries.

\section*{Acknowledgments}

This work was supported in part by the National Natural Science Foundation of China (Project No. 42230102, 42071355, 41871291), and the National Key Research and Development Program of China (Project No. 2022YFF0904400).

\bibliographystyle{elsarticle-harv}
\bibliography{StructureFacade}
\end{document}